\documentclass[letterpaper, 10 pt, conference]{ieeeconf}  

\IEEEoverridecommandlockouts                              

\usepackage{graphics} 
\usepackage{epsfig} 
\usepackage{mathptmx} 
\usepackage{times} 
\usepackage{amsmath} 
\usepackage{amssymb}  
\usepackage{lipsum} 
\usepackage{hyperref}
\usepackage{array}
\usepackage{cite}
\usepackage{xcolor}

\newcommand{\edit}[1]{{#1}}

\title{\LARGE \bf
 Exploring Behavior Discovery Methods for Heterogeneous Swarms of Limited-Capability Robots
}

\author{Connor Mattson$^{1}$, Jeremy C. Clark$^{1}$, and Daniel S. Brown$^{1}$%
\thanks{$^{1}$Kalhert School of Computing,
        University of Utah, Salt Lake City, USA
        {\tt\small c.mattson@utah.edu, j.c.clark@utah.edu, daniel.s.brown@utah.edu}}%
        }

\begin{document}

\maketitle
\thispagestyle{empty}
\pagestyle{empty}

\begin{abstract}
We study the problem of determining the emergent behaviors that are possible given a functionally heterogeneous swarm of robots with limited capabilities. Prior work has considered behavior search for homogeneous swarms and proposed the use of novelty search over either a hand-specified or learned behavior space followed by clustering to return a taxonomy of emergent behaviors to the user. In this paper, we seek to better understand the role of novelty search and the efficacy of using clustering to discover novel emergent behaviors. Through a large set of experiments and ablations, we analyze the effect of representations, evolutionary search, and various clustering methods in the search for novel behaviors in a heterogeneous swarm. Our results indicate that prior methods fail to discover many interesting behaviors and that an iterative human-in-the-loop discovery process discovers more behaviors than random search, swarm chemistry, and automated behavior discovery. The combined discoveries of our experiments uncover 23 emergent behaviors, 18 of which are novel discoveries. To the best of our knowledge, these are the first known emergent behaviors for heterogeneous swarms of computation-free agents. Videos, code, and appendix are available at the project website.\footnote[2]{\href{https://sites.google.com/view/heterogeneous-bd-methods}{https://sites.google.com/view/heterogeneous-bd-methods}.}

\end{abstract}

\section{INTRODUCTION}
One of the fundamental problems in swarm robotics is to design controllers that result in a \textit{specific} desired emergent behavior~\cite{bayindir2007review,dias2021swarm}. For example, prior work in evolutionary swarm robotics has successfully discovered controllers for aggregation~\cite{gauci2014evolving, trianni2003evolving}, shepherding~\cite{ozdemir2017shepherding}, object clustering~\cite{gauci2014clustering}, coverage~\cite{ozdemir2019spatial}, foraging~\cite{johnson2016evolving}, formation design~\cite{StolfiAutonomous, sperati2011self, st2018circle}, and collision optimization~\cite{MirhosseiniAdaptive}. However, much less work has considered the equally important question of \textit{what emergent behaviors are possible} given a swarm of robots with specific capabilities.
\edit{Prior work has categorized heterogeneous robots into two classes, structural and functional heterogeneity \cite{bae2019heuristic}. Our work focuses on functionally heterogeneous robots, where the functions of robots may differ, but structurally all robots have the same embodiment.}

\edit{Swarms of robots that act as a direct response to observations, without the need to compute any information, are known as \textit{computation-free} \cite{gauci2014self}.} Prior work on emergent behavior discovery has considered swarms of computation-free agents~\cite{brown2018discovery,mattson2023leveraging} and showed that all previously known emergent behaviors from single, line-of-sight sensor robots could be automatically discovered alongside previously undiscovered behaviors. While exciting, prior work only considers homogeneous swarms, limiting the types of interactions available when searching for swarm behaviors.

We seek to explore the richer set of emergent behaviors that result from heterogeneous swarms. We focus on swarms where each agent follows one of two different controllers (where a controller defines an agent's ``behavior type"). We compare several design choices, including type-aware vs. type-agnostic representations, hand-crafted vs. learned representations, different clustering algorithms, and human-in-the-loop vs. automated behavior discovery methods. 

We find that heterogeneous swarms of simple, computation-free robots with a single line-of-sight sensor lead to a rich set of 23 emergent behaviors (see Fig.~\ref{fig:discovered}). We find evidence that emergent behavior discovery via novelty search is less sensitive to the type of clustering algorithm used than to the representation---hierarchical, k-medoids, and spectral clustering methods all perform comparably, but using a pretrained ResNet embedding performs significantly worse than representations that are hand-crafted or learned specifically for swarm behavior search. We also find evidence that the dominant paradigm in prior work---novelty search followed by clustering---fails to discover many interesting emergent behaviors. Furthermore, we find evidence that relying on clustering to produce a taxonomy of emergent behaviors often leads to many random and uninteresting behaviors. Consequently, we find that sometimes random search outperforms novelty search. 

Motivated by these findings, we propose a new approach for emergent behavior discovery that combines novelty search with a human-in-the-loop. This approach allows us to leverage novelty search's ability to efficiently explore high-dimensional spaces while avoiding the loss of many rare behaviors by periodically showing the human the most novel behaviors found so far, rather than clustering all discovered behaviors at the end of novelty search. Our experiments show that human-in-the-loop novelty search outperforms purely automated behavior discovery by 91.4\%, random search by 38.8\%, and a pure human-guided search based on Swarm Chemistry~\cite{sayama2009swarm} by 28.15\%.

\section{PRIOR WORK}

We focus on computation-free robot swarms as originally proposed by Gauci et al.~\cite{gauci2014self}. Most prior work on computation-free swarms has only considered evolving one specific desired behavior, such as aggregation~\cite{gauci2014evolving} or area coverage~\cite{ozdemir2019spatial}. By contrast, the goal of this paper is to automatically discover the set of possible collective behaviors for swarms of mobile robots.

We analyze the use of novelty search to explore a diverse set of emergent swarm behaviors. Prior work using novelty search has focused on approaches that are designed to work with creative image and art generation or high-capability single-agent systems~\cite{gomes2014systematic,meyerson2016learning,gomes2013generic,mouret2012encouraging,nguyen2015innovation,liapis2021transforming,grillotti2022unsupervised}. 
In our work, we use novelty search to explore the space of heterogeneous swarm behaviors. Diversity is often studied in evolutionary robotics~\cite{mouret2012encouraging} and has been shown to enable an agent to search for multiple ways to accomplish a specific task~\cite{engebraaten2018evolving}.
By contrast, our work seeks to explore and categorize behavioral diversity in swarms of robots. 

Prior work on emergent behavior discovery has only considered homogeneous swarms of robots~\cite{brown2018discovery,mattson2023leveraging}. In contrast, we  analyze behavior discovery for heterogeneous robot swarms. We evaluate the use of hand-crafted feature representations~\cite{brown2018discovery} and trained embedding networks~\cite{mattson2023leveraging} (both designed for homogeneous swarms) on heterogeneous swarms and contribute augmentations to these approaches that discover more emergent behaviors. We compare these methods to Swarm Chemistry~\cite{sayama2009swarm}, which uses a human-in-the-loop to discover novel heterogeneous behaviors.
Our results show that a combination of novelty search and human-in-the-loop search results in the best performance.


\section{PROBLEM FORMULATION}

For a robot agent with sensing ($S$), memory ($M$), and actuation ($A$) capabilities, we define a robot's capability model as the three-tuple $\mathcal{C} = \langle S, M, A \rangle$. Let $U(\mathcal{C})$ represent the space of controllers that can be generated from the capability model $\mathcal{C}$.

We assign to each robot a \textit{behavior type} (for brevity, referred to as \textit{type}), that defines its controller. We define a heterogeneous swarm as a swarm of $N$ robots each associated with one of $M$ behavior types (where  $N\geq M>1$) and a corresponding controller: $c_i \in U(\mathcal{C})$, for $i=1,\ldots,M$. For a swarm of $M$ behavior types, the space of possible controllers is $U(\mathcal{C})^M$. We denote the space of possible partitions of $N$ robots into $M$ behavior types as
$\Delta^M_N$ and a specific \textit{population ratio} within the space of partitions as $\eta \in \Delta^M_N$.
Thus,
the space of all possible heterogeneous swarm configurations under $M$ behavior types is
\begin{equation}
    \label{eq:HM-swarm-configurations}
    \mathbb{H}_M = U(\mathcal{C})^M \times \Delta^M_N.
\end{equation}




Our work seeks to address the \textit{Heterogeneous Behavior Discovery Problem}: Given a robot capability model and a predetermined number of behavior types, $M$, what is the complete set of heterogeneous collective behaviors that can emerge from this multi-robot system?

\edit{In this paper, we consider the smallest heterogeneous search space where swarms contain only 2 behavior types. In the following section, we propose an approach for representing $\mathbb{H}_2$ as a set of decision variables that can be evolved and discuss methods for obtaining the set of possible emergent behaviors.}

\section{METHOD}

Automated behavior discovery for robot swarms has been shown to effectively work on limited-capability homogeneous agents by representing swarm behaviors using hand-crafted metrics~\cite{brown2018discovery} and learned embeddings \cite{mattson2023leveraging}. In this section, we discuss our proposed methods for extending and generalizing prior work in the context of discovering emergent behaviors in heterogeneous swarms. 

\subsection{Swarm Configuration}

\begin{figure}[t]
     \centering
     \includegraphics[width=0.5\linewidth]{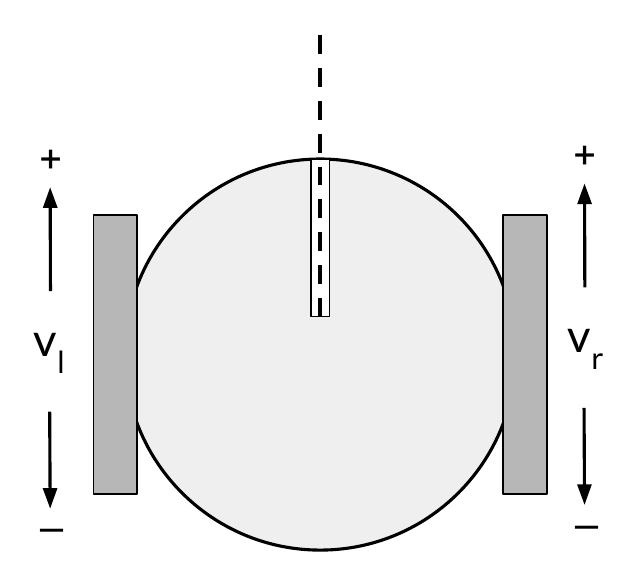}
     \caption{A computation-free capability model, where a sensor positioned in the forward orientation is a binary line-of-sight sensor. The robot can actuate both the right and left wheels with velocities $v_l$ and $v_r$.}
    \label{fig:capability-model}
\end{figure}

Following prior work~\cite{brown2018discovery, mattson2023leveraging}, we use a robot capability model that is computation-free, where decisions are made using only \edit{a single bit of input from a binary sensor}. As shown in Fig.~\ref{fig:capability-model}, the robot model controls two actuators, corresponding to the left and right wheels, by commanding a desired velocity to each actuator. This robot is capable of sensing other agents using an infinite-distance line-of-sight sensor, positioned in the forward position relative to the robot chassis. This sensor outputs a binary signal, $\{0, 1\}$, which is used to actuate the robot according to a single conditional branch, commanding one pair of velocities to the wheels when the sensor signal is $0$ and a different pair of velocities to the wheels when the sensor signal is $1$. Under this capability model and the 2-branch conditional decision scheme, all actions taken by this agent over its entire lifetime can be represented by 4 real-valued velocities, 2 for each binary state.

\begin{figure}[t]
     \centering
     \includegraphics[width=\linewidth]{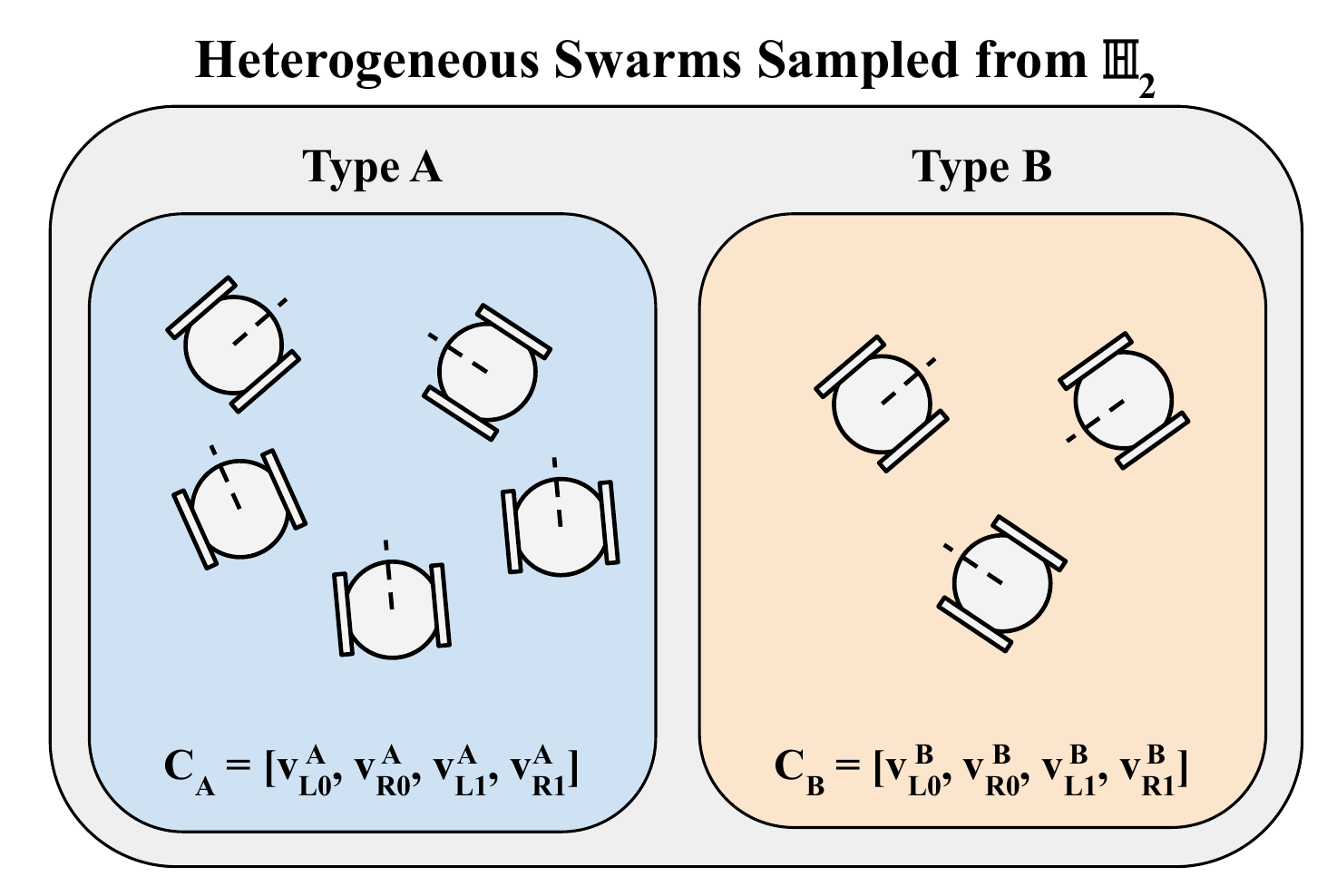}
     \caption{For a \edit{functionally} heterogeneous swarm comprised of $N$ binary-controlled differential drive robots and two behavior types, the complete set of swarm configurations can be sampled from a vector of 9 real-valued numbers (8 values for the controllers used by behavior types A and B, and 1 for the population ratio, $\eta$). For the example shown, $N = 8$ and $\eta = \frac{5}{8}$.}
    \label{fig:h2-sampling}
\end{figure}

To uncover the set of possible collective behaviors, we examine sampling and search methods over the space of heterogeneous swarm configurations $\mathbb{H}_2$. For a 2-type heterogeneous swarm, we formulate a configuration space from the the selection of two controllers and the assignment of some fraction of the agents to behavior type $A$ based on a \textit{population ratio}, $\eta \in (0, 1)$, assigning the remaining $(1 - \eta)$ fraction of agents to behavior type $B$. Formally, a swarm consisting of 2 behavior types can be sampled from the space of decision variables formed by

\begin{equation}
    \label{eq:decision-space}
    \mathbb{H}_2 = [v^{A}_{l0}, v^{A}_{r0}, v^{A}_{l1}, v^{A}_{r1}, v^{B}_{l0}, v^{B}_{r0}, v^{B}_{l1}, v^{B}_{r1}, \eta],
\end{equation}
where, for types $\{A, B\}$, the velocities commanded to the left ($l$) and right ($r$) wheels are $v_{\{l, r\}0}$ when the binary sensor detects nothing and $v_{\{l, r\}1}$ when the sensor detects another agent (Fig. \ref{fig:h2-sampling}). Velocities are restricted to the range $[-1, 1]$, resulting in the continuous search space $[-1, 1]^8 \times (0, 1)$.

\subsection{Behavioral Representation}
\label{sec:b-representation}

\begin{figure}[t]
     \centering
     \includegraphics[width=\linewidth]{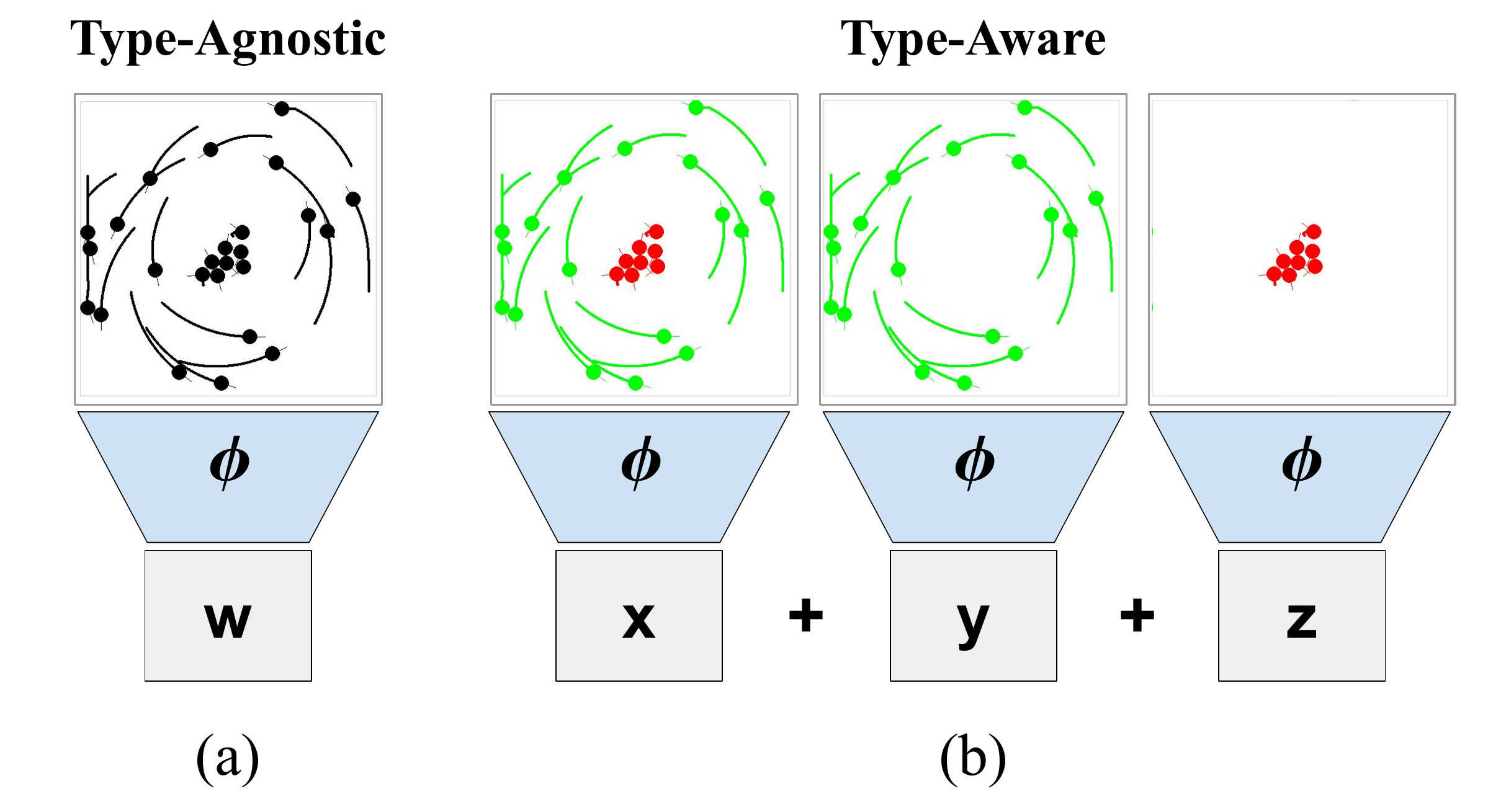}
     \caption{Representation Classes used in experimentation. (a) Type-agnostic Representations encode information, via the encoder $\phi$, by considering the entirety of the swarm as one type, where there is no explicit learning from type information. (b) Type-aware representations encode information about the swarm as a whole, but also explicitly encode data using privileged knowledge of each type (shown in red and green), \edit{which reflect type-aware reprsentations of the swarm when concatenated together. Here, w, x, y, and z are real vectors in the output dimensions of the encoder $\phi$.}} 
    \label{fig:representations}
\end{figure}

The complex nature of agent interaction and locomotion in robot swarms necessitates systematic low-dimensional representations of high-level behavior. 
In this work, we consider two classes of behavior representations, \textit{type-agnostic} representations and \textit{type-aware} representations (Fig. \ref{fig:representations}). For type-agnostic representations, there is no explicit distinction provided between the two behavior types. Prior work in automated behavior discovery~\cite{brown2018discovery, mattson2023leveraging}, addressed behavior discovery for homogeneous swarms, mitigating any need to consider behavior occurring at the sub-swarm level. Introducing multiple robot types to an environment allows us to examine representations that encode additional information about type-level behavior. We study the effect of type-aware representations, where behaviors are calculated or encoded using privileged knowledge of the locality and behavior that exists within each type. 

We examine the use of both type-agnostic and type-aware versions of the following representations in the search for new collective emergent behaviors:

\subsubsection{Hand-Crafted Behavior Features~\cite{brown2018discovery}}
A vector in $\mathbb{R}^5$ \edit{calculated from 5 feature equations: average speed, angular momentum, radial variance, scatter, and group rotation, as described in \cite{brown2014balancing}}. For type-aware representations, the metrics are calculated for each of the two types as well as the entire swarm. These vectors are then concatenated to form a feature vector in $\mathbb{R}^{15}$.

\subsubsection{Learned Behavior Embedding~\cite{mattson2023leveraging}}
A pretrained Convolutional Neural Network (CNN) trained via self-supervised learning and human labeling to embed homogeneous swarm trajectories into a latent vector in $\mathbb{R}^5$. For type-agnostic representations, swarm trajectories are rendered in simulation with all agents displayed with the same color (Fig. \ref{fig:representations}a), collapsed into a single channel image of size 50x50, and then embedded into $\mathbb{R}^5$. Type-aware representations have specifically colored behavior types, where red and green agents can be extracted into the red and green image channels and embedded separately into the network (which was trained on single-channel images). The two channel embeddings are then concatenated to the embedding of the type-agnostic embedding to form a vector in $\mathbb{R}^{15}$ (Fig. \ref{fig:representations}b). 


\subsubsection{Pretrained ResNet18~\cite{he2016deep}}
We also consider the use of an off-the-shelf feature representation model that has not been trained on robot swarms before. We examine the use of a pretrained ResNet18~\cite{he2016deep}, a CNN trained to classify images on 1,000 classes, as a method of feature extraction for our swarm trajectories. Swarm trajectories are resized to 256x256 and embedded into the final convolutional layer, which is flattened to form a behavior representation in $\mathbb{R}^{512}$. For ResNet, our type-agnostic representation is a \edit{single-channel greyscale trajectory (Fig. \ref{fig:representations}a) copied into three image channels to match the input size of ResNet} and the type-aware representation is the 3-channel colored trajectories. 



\subsection{Taxonomy Search and Formulation}
Following prior methods by Brown et al.~\cite{brown2018discovery} and Mattson et al.~\cite{mattson2023leveraging}, we explore and evolve a diverse set of emergent swarm behaviors using Novelty Search~\cite{lehman2011abandoning}, an evolutionary algorithm that rewards exploration and diversity in phenotypes. Given an initial population of sampled controllers from $\mathbb{H}_2$, evolutionary priority for crossover and mutation is given to controllers in the following generation based off the novelty of the associated behavior representation~\cite{lehman2011abandoning}:
\begin{equation}
    \label{eq:novelty-score}
    Novelty(v) = \frac{1}{p} \sum_{i=0}^p dist(v, \mathbb{A}_i),
\end{equation}
where the distance between the phenotype parameter, $v$, and the $p$-nearest-neighbors in the archive of previously explored phenotypes, $\mathbb{A}$, is averaged to compute a positive real-valued novelty score. The goal of each generation is to produce outcomes that are novel when compared to the phenotypes of previous generations. All of the evolved behaviors form an archive $\mathbb{A}$, which is used in prior work as the dataset for a k-medoids clustering to extract a taxonomy of $k$ emergent behaviors~\cite{brown2018discovery,mattson2023leveraging}. While prior work only considers the use of k-medoids to formulate a taxonomy, we also explore the use of hierarchical and spectral data clustering algorithms.

\begin{figure*}[h]
     \centering
     \includegraphics[width=\linewidth]{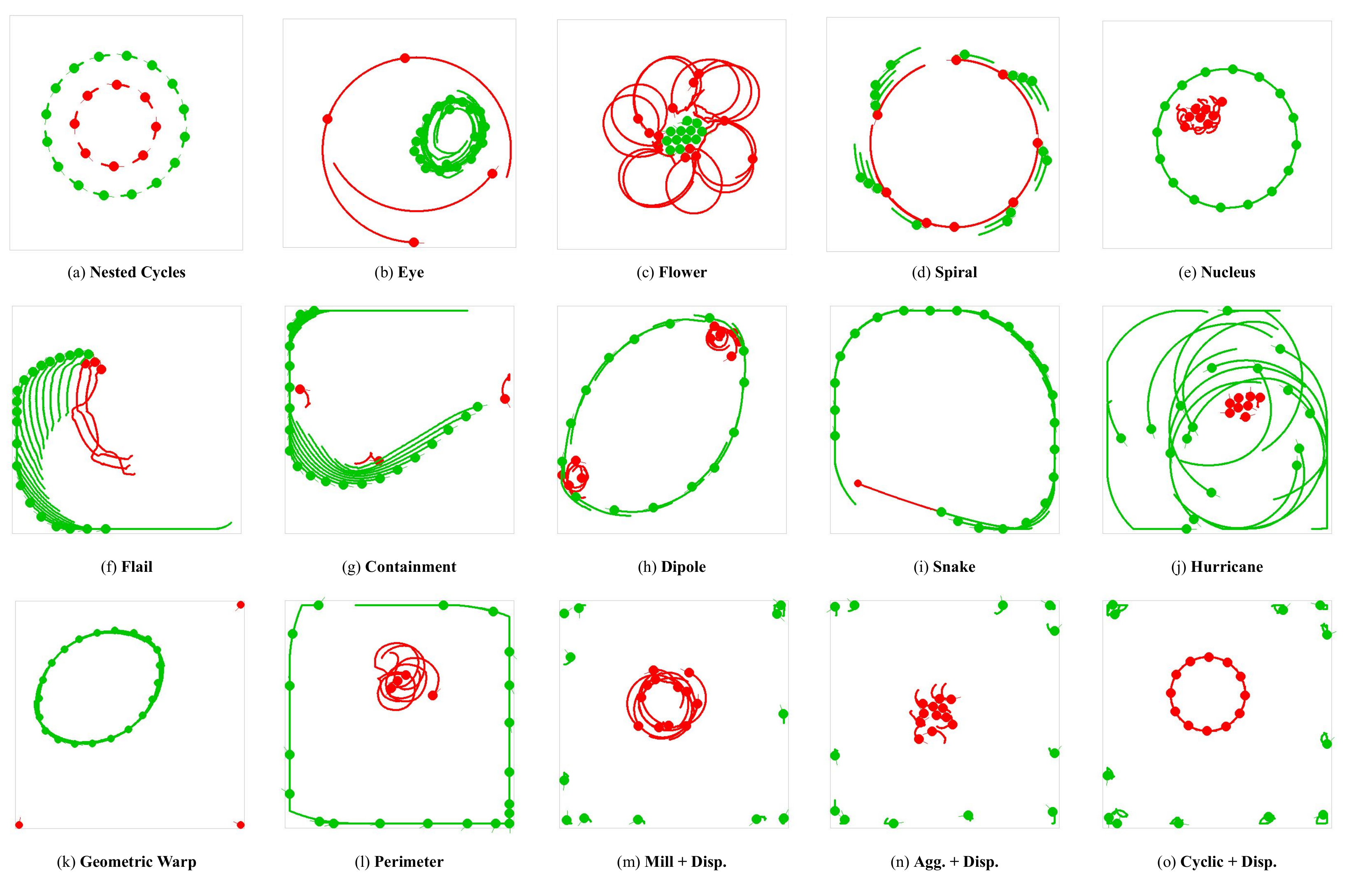}
     \caption{15 of 23 heterogeneous behaviors that were discovered throughout our study. Robots are displayed as red (Type 1) and green (Type 2) circles with traces of their recent actions shown as a curve connected to each agent. Videos of these behaviors are available on our website at \href{https://sites.google.com/view/heterogeneous-bd-methods}{https://sites.google.com/view/heterogeneous-bd-methods}.}
    \label{fig:discovered}
\end{figure*}

\section{EXPERIMENTS}

To analyze the methods described in the previous section, we conduct a series of experiments in simulation to answer three research questions: 
(1) Does the representation type and clustering method matter when automatically discovering new behaviors? (Sec. \ref{sec:experiments-represention-discovery}); 
(2) Does Novelty Search \cite{lehman2011abandoning} outperform a set of randomly sampled controllers? (Sec. \ref{sec:experiments-Ns-vs-Rand});
(3) Can we utilize a human-in-the-loop paired with novelty search to improve the diversity of our behavior taxonomy? (Sec. \ref{sec:experiments-hil})


We conduct our experiments in 2D simulation using a custom Pygame~\cite{pygame} simulator developed in our prior work~\cite{mattson2023leveraging}. All results represent an average over 3 trials with seeds \{0, 1, 2\}. Additional experiment parameters are included in the appendix.

\subsection{Discovered Heterogeneous Behaviors}
Throughout our experiments, we find a rich set of 23 emergent heterogeneous behaviors, 18 of which, to the best of our knowledge, have never been discovered for agents of the computation-free capability model (Fig. \ref{fig:discovered}). 
\edit{Following prior work \cite{mattson2023leveraging}, our experimental results reflect behaviors that are designated as subjectively interesting and distinct by a human overseer. With this in mind, we justify distinctions between all emergent behaviors in Appendix \ref{appendix:discovered-behaviors} and keep the definitions of behaviors consistent across all experiments.}

The behaviors we discover validate the benefits of heterogeneity by highlighting unique interactions that can exist between types of robots with differing controllers. For example, in the flower behavior (Fig. \ref{fig:discovered}c), we observe that green agents and red agents are both performing aggregation but red agents have a controller with a large turning radius, causing them to deviate from and then return to the group, filtering the green agents into the center. We also see the emergence of some adversarial behaviors, including Containment (Fig. \ref{fig:discovered}g), where green agents attempt to surround a group of escaping red agents, and Snake (Fig. \ref{fig:discovered}i), where a line of green agents chases one or more red agents. 

We find that most heterogeneous behaviors for 2 robot types can be expressed as a combination of 2 emergent homogeneous behaviors. In some cases, such as Fig. \ref{fig:discovered}[l-o], this distinction is clear as the two robot types are isolated from each other in the environment. However, in some cases, such as Flail (Fig. \ref{fig:discovered}f), Dipole (Fig. \ref{fig:discovered}h),
and Spiral (Fig. \ref{fig:discovered}d) the interactions between the robot types likely would not have been intuitive to a swarm engineer, further validating the exciting potential for behavior discovery methods to help uncover exciting new robot interactions.

\begin{figure*}[h]
     \centering
     \includegraphics[width=0.95\linewidth]{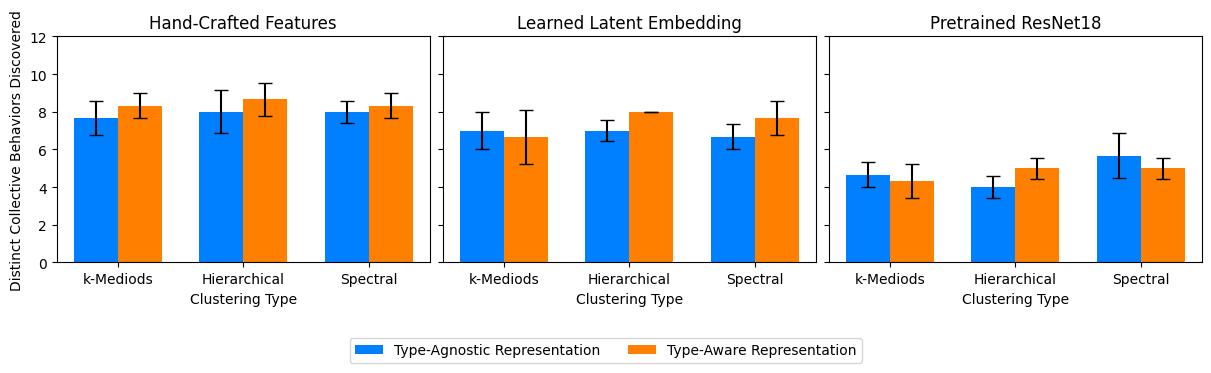}
     \caption{The number of distinct emergent behaviors uncovered during automatic search across 3 different behavior representations and aware-agnostic trials for 3 types of clustering algorithms. Error bars shown reflect the standard error across 3 trials.}
    \label{fig:results-automated}
\end{figure*}

\begin{table*}[]
\caption{Number of Distinct Emergent Behaviors Discovered for several clustering approaches and representation models. 
}
\begin{tabular}{lccccccccc}
\hline
& \multicolumn{3}{c}{Hand-Crafted Features} & \multicolumn{3}{c}{Learned Embedding}     & \multicolumn{3}{c}{Pretrained ResNet18} \\ \hline
& k-Medoids    & Hierarchical & Spectral    & k-Medoids   & Hierarchical & Spectral    & k-Medoids   & Hierarchical & Spectral   \\ \hline
Agnostic NS     & 7.67$\pm$0.88            & 8.0$\pm$1.15          & 8.0$\pm$0.57          & 7.0$\pm$1.00           & 7.0$\pm$0.57            & 6.67$\pm$0.66           & 4.67$\pm$0.66           & 4.0$\pm$0.57            & 5.67$\pm$1.19          \\
Aware NS        &  \textbf{8.33$\pm$0.66}  & \textbf{8.67$\pm$0.88}  &  \textbf{8.33$\pm$0.66} & 6.67$\pm$1.45  & \textbf{8.0$\pm$0.0} & \textbf{7.67$\pm$0.88} & 4.33$\pm$0.88 & 5.0$\pm$0.57  & 5.0$\pm$0.57          \\
Agnostic Random & 6.33$\pm$0.33  & 7.0$\pm$0.57  &  7.0$\pm$0.57 & 6.67$\pm$0.66 & 6.33$\pm$0.33 & 7.0$\pm$1.15 & \textbf{8.67$\pm$0.33} & 7.33$\pm$0.88 & 7.0$\pm$1.00 \\
Aware Random & 6.0$\pm$0.57 & 7.33$\pm$0.66 & 5.67$\pm$0.33 & \textbf{8.0$\pm$0.57} & 7.0$\pm$0.00 & 7.33$\pm$0.33 & 7.0$\pm$1.00 & \textbf{7.67$\pm$0.66} & \textbf{8.33$\pm$0.33} \\ \hline
\end{tabular}
\label{tab:random-comparison}
\end{table*}

\subsection{Representation-Type and Clustering Algorithm Results}
\label{sec:experiments-represention-discovery}

In this experiment, we test whether heterogeneous behavior discovery outputs a more diverse set of behaviors when encoding type-aware information compared to type-agnostic information.
For each behavior representation, we compare Novelty Search using type-aware data and type-agnostic data. Each search is run for 50 generations with each generation simulating 100 swarms  for a total of 5000 achieved behaviors. Using these archives, we create k=20 clusters where we extract the controller whose representation is closest to the centroid of each cluster and hand-label the corresponding behavior to aggregate the total number of distinct behaviors discovered.

We compare the number of distinct collective behaviors discovered for each combination of representation and clustering approach (Fig.~\ref{fig:results-automated}). Our analysis indicates that there is no significant evidence to suggest a clear advantage in choosing either type-agnostic or type-aware representations. Our results also shows that the learned embedding and hand-crafted representations of homogeneous behaviors both outperform an off-the-shelf ResNet18. Finally, we found no clear evidence suggesting that one type of clustering is preferred for behavior discovery over another.

\subsection{Novelty Search Versus Random Sampling Results}
\label{sec:experiments-Ns-vs-Rand}

To further analyze prior automated behavior discovery methods, we consider whether novelty search is necessary to evolve a diverse archive of behaviors for use in clustering. Prior work \cite{brown2018discovery, mattson2023leveraging} has used Novelty Search as a means for behavior discovery and Mattson et al.~\cite{mattson2023leveraging} showed that novelty search and clustering over a learned representation space was an improvement compared to randomly sampling $k$ controllers from the search space. In this experiment, we propose a stronger baseline, where we generate the archive $\mathbb{A}$ with 5000 random samples from $\mathbb{H}_2$. Using the randomly sampled archive, we employ the same clustering strategies to extract a set of emergent behaviors.

In Table \ref{tab:random-comparison}, we compare the number of distinct emergent behaviors obtained when using novelty search to form our novelty archive and when the archive is formed from random sampling. We find that using Brown et al.'s~\cite{brown2018discovery} hand-crafted features for type-aware novelty search outperforms both variations of random sampling, leading to 31.59\% more behaviors being discovered compared to type-agnostic random (which performed best out of the two random trials) for k-medoids, 18.28\% more than type-aware random for hierarchical clustering, and 19\% more than type-agnostic random for spectral clustering.

For Mattson et al.'s~\cite{mattson2023leveraging} learned embedding, the type-aware novelty search outperforms type-aware Random by 14.2\% for hierarchical and 4.6\% for spectral clustering. However, the type-aware random k-medoids outperforms the novelty search approach for the learned behavior embedding. For the ResNet18, we find that random sampling always outperforms novelty search. 

Our results indicate that the performance of random search with ResNet18 representations is very similar to the performance of novelty search and clustering for the other representations. We hypothesize that the reason Novelty Search only produced, on average, minimal gains over randomly constructed archives is that uninteresting behaviors were much more likely to be present in the centroids of clustering than random sampling (4.95\% increase in uninteresting behaviors for Hand-Crafted Features, 39.51\% increase for the learned embedding, and 147.91\% increase for ResNet, see appendix). Specifically, this significant increase in random behaviors indicates that novelty search spent a lot of time exploring behaviors that are largely uninteresting to the human, but were novel at the representation-level. This explains why ResNet18, which has no prior representation training on swarm behaviors, would explore far more uninteresting behaviors than random sampling alone. If the archives generated from novelty search contain dense regions of uninteresting datapoints, it is more likely that using clustering for taxonomy formulation will return a large number of uninteresting behaviors.

We believe that the marginal gains from clustering the novelty archive do not nullify the potential benefits of novelty search. Rather, these results lead to a natural follow-up question: how can we better utilize novelty search to form a more representative taxonomy of what was explored? In the following section, we consider this question.

\subsection{Human-in-the-loop Improvement}
\label{sec:experiments-hil}

Novelty search computes a novelty score (Eq. \ref{eq:novelty-score}) for each evolved behavior at the end of every generation. In this experiment, we propose a novel Human-in-the-Loop Novelty Search (HIL-NS) approach that strategically queries a human based on these novelty scores as a replacement for taxonomy extraction via clustering. 
Using the same evolutionary strategy as the previous experiments, we present a human with the 3 most novel behaviors at the end of each generation of novelty search. The human may choose to add any of these behaviors to an aggregated taxonomy of interesting behaviors they have discovered. At the end of 50 generations, the returned taxonomy is the aggregated set of the behaviors that the human saved during the search. To the best of our knowledge, we are the first to consider using a combination of novelty search and human-in-the-loop feedback to construct a novel taxonomy for swarm behaviors. \edit{When applied to our swarm, we find that HIL-NS extracts 16.66 emergent behaviors, on average, across 3 trials.}

\begin{figure}[t]
     \centering
     \includegraphics[width=0.9\linewidth]{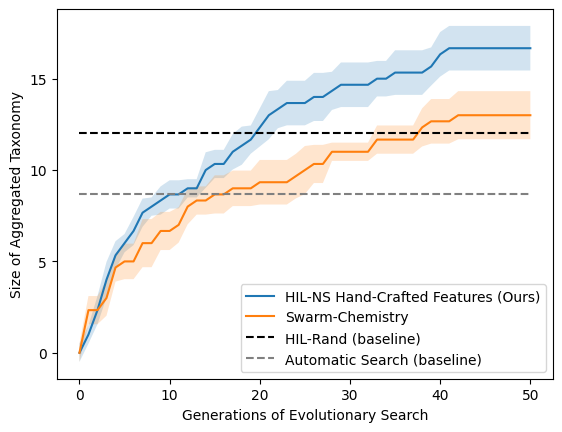}
     \caption{A comparison of disinct emergent behaviors discovered for HIL-NS, a human-in-the-loop implementation of novelty search~\cite{lehman2011abandoning} that aggregates a taxonomy of emergent behaviors by choosing specific queries for the user to label, and Swarm Chemistry~\cite{sayama2009swarm}, the state-of-the-art approach to human-guided behavior evolution. These evolutionary approaches are compared to human-saved behaviors collected from 150 random samples (HIL-Rand), and the best result from Automatic Search described in Sec. \ref{sec:experiments-represention-discovery}.}
    \label{fig:HIL-NS-vs-Sayama}
\end{figure}

We compare this method to Swarm Chemistry~\cite{sayama2009swarm}, a prior approach that uses a human-in-the-loop as a fitness function in an evolutionary walk through of the behavior space. In Swarm Chemistry, the human selects the behaviors which they deem to be the most interesting over a series of generations. The associated controllers are then randomly mutated and/or combined to produce the next generation (more details can be found in our Appendix). Though Swarm Chemistry is effective at discovering new behaviors, we find that over 50 generations of search, HIL-NS discovers 28.15\% more emergent behaviors on average and 38.8\% more than what the human found interesting from 150 randomly sampled behaviors (HIL-Rand) (Fig \ref{fig:HIL-NS-vs-Sayama}). \edit{When compared to the best result from Automatic Search (Table \ref{tab:random-comparison}), HIL-NS finds 91.4\% more behaviors on average.}   

\section{CONCLUSION}
This paper considers the behavior discovery problem where, given a robot's capability model, the complete set of emergent swarm behaviors can be efficiently discovered. We provide an analysis of prior swarm behavior discovery methods applied to heterogeneous swarms of limited-capability robots and find that a human-in-the-loop novelty search approach outperforms random search, fully-automatic behavior discovery, and Swarm Chemistry by 38.8\%, 91.4\%, and 28.15\%, respectively. We show that local interactions between heterogeneous robot types can lead to 23 distinct behavioral patterns, 18 of which are novel discoveries for robots of the computation-free capability model.


Our results highlight the diminishing effectiveness of combining novelty search and clustering together as the dimensionality of the search space and rarity of interesting behaviors increases. We hypothesize that this is because novelty search will continue to pursue areas of the behavior space that produce random behaviors that, while uninteresting and/or incoherent to a human, appear distinct in terms of representation features. To address this problem, we propose a novel approach for using feedback from the human during novelty search to improve behavior discovery.

Future work should examine these approaches on more complicated robots and in swarms containing more than 2 types of robots, where we believe even more interesting swarm behaviors lie undiscovered. \edit{It should be noted that our work does not include a full user study---our human experiments reflect the participation and expertise of the authors. An interesting area of future work is to run a user study with non-experts to explore how a general audience perceives emergent behaviors and compare discovered behaviors across different users.} In addition, future work should explore how humans can provide input that refines novelty search or behavior representations so that behaviors uncovered during search further align with the human's beliefs about which behaviors are novel. 








\bibliographystyle{plain}
\bibliography{main}

\begin{table*}[]
\centering
\caption{Frequency of Behavior Extraction for Novelty Search (NS) and Random Search (Rand.) accross Hand-Crafted (HC), Learned Latent Embedding (LL) and ResNet18 (RN) representations. The dashes (---) denote behaviors that were not discovered.}
\begin{tabular}{lcccccccccccc}                                                                                                                              \\ \cline{2-13} 
\multicolumn{1}{l|}{}                            & \multicolumn{2}{c|}{HC Agnostic}     & \multicolumn{2}{c|}{HC Aware}        & \multicolumn{2}{c|}{LL Agnostic}     & \multicolumn{2}{c|}{LL Aware}        & \multicolumn{2}{c|}{RN Agnostic}     & \multicolumn{2}{c|}{RN Aware}        \\
\multicolumn{1}{l|}{Behavior}                    & NS     & \multicolumn{1}{c|}{Rand.}  & NS     & \multicolumn{1}{c|}{Rand.}  & NS     & \multicolumn{1}{c|}{Rand.}  & NS     & \multicolumn{1}{c|}{Rand.}  & NS     & \multicolumn{1}{c|}{Rand.}  & NS     & \multicolumn{1}{c|}{Rand.}  \\ \hline
\multicolumn{1}{|l|}{Random}                     & 32.2\% & \multicolumn{1}{c|}{38.9\%} & 38.3\% & \multicolumn{1}{c|}{28.3\%} & 48.3\% & \multicolumn{1}{c|}{33.9\%} & 47.8\% & \multicolumn{1}{c|}{35.0\%} & 71.7\% & \multicolumn{1}{c|}{26.7\%} & 60.6\% & \multicolumn{1}{c|}{26.7\%} \\
\multicolumn{1}{|l|}{Cyclic Pursuit}             & ---  & \multicolumn{1}{c|}{---}  & ---  & \multicolumn{1}{c|}{---}  & ---  & \multicolumn{1}{c|}{---}  & 0.6\%  & \multicolumn{1}{c|}{---}  & 0.6\%  & \multicolumn{1}{c|}{---}  & ---  & \multicolumn{1}{c|}{---}  \\
\multicolumn{1}{|l|}{Milling}                    & 1.7\%  & \multicolumn{1}{c|}{1.7\%}  & 6.7\%  & \multicolumn{1}{c|}{0.6\%}  & 2.2\%  & \multicolumn{1}{c|}{0.6\%}  & 3.9\%  & \multicolumn{1}{c|}{0.6\%}  & 1.1\%  & \multicolumn{1}{c|}{2.8\%}  & 1.1\%  & \multicolumn{1}{c|}{2.8\%}  \\
\multicolumn{1}{|l|}{Aggregation}                & 1.1\%  & \multicolumn{1}{c|}{6.1\%}  & 0.6\%  & \multicolumn{1}{c|}{5.6\%}  & 3.9\%  & \multicolumn{1}{c|}{5.6\%}  & 4.4\%  & \multicolumn{1}{c|}{10.6\%} & ---  & \multicolumn{1}{c|}{8.9\%}  & 2.8\%  & \multicolumn{1}{c|}{11.7\%} \\
\multicolumn{1}{|l|}{Dispersal}                  & 6.1\%  & \multicolumn{1}{c|}{13.3\%} & 5.6\%  & \multicolumn{1}{c|}{2---} & 11.1\% & \multicolumn{1}{c|}{16.7\%} & 1.7\%  & \multicolumn{1}{c|}{9.4\%}  & 5.0\%  & \multicolumn{1}{c|}{15.6\%} & 1--- & \multicolumn{1}{c|}{7.2\%}  \\
\multicolumn{1}{|l|}{Wall-Following}             & ---  & \multicolumn{1}{c|}{---}  & ---  & \multicolumn{1}{c|}{---}  & ---  & \multicolumn{1}{c|}{1.1\%}  & 0.6\%  & \multicolumn{1}{c|}{---}  & 1.1\%  & \multicolumn{1}{c|}{2.8\%}  & 0.6\%  & \multicolumn{1}{c|}{---}  \\
\multicolumn{1}{|l|}{Nested Cycles}              & 2.8\%  & \multicolumn{1}{c|}{3.3\%}  & 0.6\%  & \multicolumn{1}{c|}{2.2\%}  & 1.1\%  & \multicolumn{1}{c|}{---}  & 1.7\%  & \multicolumn{1}{c|}{2.2\%}  & 0.6\%  & \multicolumn{1}{c|}{6.1\%}  & 1.7\%  & \multicolumn{1}{c|}{2.8\%}  \\
\multicolumn{1}{|l|}{Containment}                & 16.7\% & \multicolumn{1}{c|}{8.3\%}  & 6.7\%  & \multicolumn{1}{c|}{6.7\%}  & 7.2\%  & \multicolumn{1}{c|}{9.4\%}  & 1.7\%  & \multicolumn{1}{c|}{6.7\%}  & 2.8\%  & \multicolumn{1}{c|}{3.9\%}  & 4.4\%  & \multicolumn{1}{c|}{8.3\%}  \\
\multicolumn{1}{|l|}{Spiral}                     & 0.6\%  & \multicolumn{1}{c|}{---}  & 2.2\%  & \multicolumn{1}{c|}{5.6\%}  & 1.1\%  & \multicolumn{1}{c|}{---}  & 1.1\%  & \multicolumn{1}{c|}{1.1\%}  & ---  & \multicolumn{1}{c|}{---}  & ---  & \multicolumn{1}{c|}{---}  \\
\multicolumn{1}{|l|}{Segments}                   & 1.7\%  & \multicolumn{1}{c|}{1.1\%}  & 1.1\%  & \multicolumn{1}{c|}{1.1\%}  & 1.7\%  & \multicolumn{1}{c|}{1.1\%}  & 4.4\%  & \multicolumn{1}{c|}{1.1\%}  & 2.2\%  & \multicolumn{1}{c|}{5.6\%}  & 4.4\%  & \multicolumn{1}{c|}{2.2\%}  \\
\multicolumn{1}{|l|}{Nucleus}                    & 6.1\%  & \multicolumn{1}{c|}{8.9\%}  & 3.9\%  & \multicolumn{1}{c|}{11.7\%} & 6.7\%  & \multicolumn{1}{c|}{7.8\%}  & 8.9\%  & \multicolumn{1}{c|}{6.1\%}  & 1.1\%  & \multicolumn{1}{c|}{13.9\%} & 3.3\%  & \multicolumn{1}{c|}{9.4\%}  \\
\multicolumn{1}{|l|}{Site Traversal}             & ---  & \multicolumn{1}{c|}{0.6\%}  & 0.6\%  & \multicolumn{1}{c|}{---}  & ---  & \multicolumn{1}{c|}{---}  & 0.6\%  & \multicolumn{1}{c|}{1.1\%}  & 0.6\%  & \multicolumn{1}{c|}{0.6\%}  & 3.3\%  & \multicolumn{1}{c|}{1.1\%}  \\
\multicolumn{1}{|l|}{Flail}                      & ---  & \multicolumn{1}{c|}{---}  & ---  & \multicolumn{1}{c|}{0.6\%}  & 0.6\%  & \multicolumn{1}{c|}{---}  & ---  & \multicolumn{1}{c|}{0.6\%}  & ---  & \multicolumn{1}{c|}{---}  & 0.6\%  & \multicolumn{1}{c|}{0.6\%}  \\
\multicolumn{1}{|l|}{Dipole}                     & ---  & \multicolumn{1}{c|}{---}  & 0.6\%  & \multicolumn{1}{c|}{---}  & ---  & \multicolumn{1}{c|}{---}  & 1.1\%  & \multicolumn{1}{c|}{0.6\%}  & ---  & \multicolumn{1}{c|}{---}  & ---  & \multicolumn{1}{c|}{---}  \\
\multicolumn{1}{|l|}{Hurricane}                  & 1.1\%  & \multicolumn{1}{c|}{3.3\%}  & 3.9\%  & \multicolumn{1}{c|}{1.7\%}  & 2.8\%  & \multicolumn{1}{c|}{5.6\%}  & 5.6\%  & \multicolumn{1}{c|}{1.1\%}  & 2.8\%  & \multicolumn{1}{c|}{1.1\%}  & ---  & \multicolumn{1}{c|}{2.2\%}  \\
\multicolumn{1}{|l|}{Snake}                      & 5.0\%  & \multicolumn{1}{c|}{2.8\%}  & 5.0\%  & \multicolumn{1}{c|}{1.1\%}  & 0.6\%  & \multicolumn{1}{c|}{1.1\%}  & 0.6\%  & \multicolumn{1}{c|}{---}  & 0.6\%  & \multicolumn{1}{c|}{2.8\%}  & 2.8\%  & \multicolumn{1}{c|}{1.7\%}  \\
\multicolumn{1}{|l|}{Milling + Dispersal}        & 5.6\%  & \multicolumn{1}{c|}{---}  & 7.8\%  & \multicolumn{1}{c|}{0.6\%}  & 3.9\%  & \multicolumn{1}{c|}{5.6\%}  & 8.3\%  & \multicolumn{1}{c|}{5.0\%}  & 2.2\%  & \multicolumn{1}{c|}{0.6\%}  & 2.2\%  & \multicolumn{1}{c|}{2.8\%}  \\
\multicolumn{1}{|l|}{Aggregation + Dispersal}    & 7.2\%  & \multicolumn{1}{c|}{10.6\%} & 5.6\%  & \multicolumn{1}{c|}{13.9\%} & 3.3\%  & \multicolumn{1}{c|}{9.4\%}  & 3.3\%  & \multicolumn{1}{c|}{15.6\%} & 5.0\%  & \multicolumn{1}{c|}{6.1\%}  & 1.7\%  & \multicolumn{1}{c|}{16.7\%} \\
\multicolumn{1}{|l|}{Cyclic Pursuit + Dispersal} & 7.2\%  & \multicolumn{1}{c|}{0.6\%}  & 6.7\%  & \multicolumn{1}{c|}{---}  & 3.3\%  & \multicolumn{1}{c|}{0.6\%}  & 0.6\%  & \multicolumn{1}{c|}{2.8\%}  & 1.1\%  & \multicolumn{1}{c|}{1.7\%}  & 0.6\%  & \multicolumn{1}{c|}{2.8\%}  \\
\multicolumn{1}{|l|}{Geometric Warp}             & 0.6\%  & \multicolumn{1}{c|}{---}  & 1.1\%  & \multicolumn{1}{c|}{---}  & ---  & \multicolumn{1}{c|}{0.6\%}  & ---  & \multicolumn{1}{c|}{---}  & 1.1\%  & \multicolumn{1}{c|}{---}  & ---  & \multicolumn{1}{c|}{---}  \\
\multicolumn{1}{|l|}{Mill Followers}             & 2.8\%  & \multicolumn{1}{c|}{0.6\%}  & 2.2\%  & \multicolumn{1}{c|}{---}  & 2.2\%  & \multicolumn{1}{c|}{0.6\%}  & 1.7\%  & \multicolumn{1}{c|}{0.6\%}  & 0.6\%  & \multicolumn{1}{c|}{0.6\%}  & ---  & \multicolumn{1}{c|}{1.1\%}  \\
\multicolumn{1}{|l|}{Perimeter}                  & ---  & \multicolumn{1}{c|}{---}  & ---  & \multicolumn{1}{c|}{---}  & ---  & \multicolumn{1}{c|}{---}  & ---  & \multicolumn{1}{c|}{---}  & ---  & \multicolumn{1}{c|}{0.6\%}  & ---  & \multicolumn{1}{c|}{---}  \\
\multicolumn{1}{|l|}{Flower}                     & 1.1\%  & \multicolumn{1}{c|}{---}  & ---  & \multicolumn{1}{c|}{0.6\%}  & ---  & \multicolumn{1}{c|}{---}  & 0.6\%  & \multicolumn{1}{c|}{---}  & ---  & \multicolumn{1}{c|}{---}  & ---  & \multicolumn{1}{c|}{---}  \\
\multicolumn{1}{|l|}{Eye}                        & 0.6\%  & \multicolumn{1}{c|}{---}  & 1.1\%  & \multicolumn{1}{c|}{---}  & ---  & \multicolumn{1}{c|}{0.6\%}  & 1.1\%  & \multicolumn{1}{c|}{---}  & ---  & \multicolumn{1}{c|}{---}  & ---  & \multicolumn{1}{c|}{---}  \\ \hline
\end{tabular}
\label{tab:appendix-freq-table}
\end{table*}
\newpage
\section*{APPENDIX}

\subsection{Experiment Details}
At the start of every simulation, each agent is initialized to a random position and orientation $(x, y, \theta)$ in an empty 500x500 units environment. All agents simultaneously begin operation at time $t=0$ and proceed to collect sensor data and actuate accordingly every timestep until the timeout horizon at $t=1200$. 

Our heterogeneous swarms consist of 24 differential drive agents (Fig. \ref{fig:capability-model}). To limit the search space for $\mathbb{H}_2$ (Eq. \ref{eq:decision-space}), we explore the discrete set of velocities $v \in \{-1.0, -0.9, ..., 0.9, 1.0\}$ and a discrete set of population ratios $\eta \in \{\frac{1}{24}, \frac{3}{24}, \frac{6}{24}, \frac{8}{24}, \frac{12}{24}\}$, resulting in $21^8 \times 5 = 1.89$x$10^{11}$ possible heterogeneous swarm configurations. Out implementation of novelty search uses a 0.15 mutation rate, 0.7 crossover rate, and p=14 (Eq. \ref{eq:novelty-score}).

\subsection{Behavioral Frequency in Automated Search}
Our results indicate that Behavior Discovery using Novelty Search did not reliably perform better for heterogeneous swarms than clustering over randomly sampled archives. We hypothesized that this had something to do with the number of random behaviors that were being uncovered in search. Table \ref{tab:appendix-freq-table} shows the frequency of the 23 behaviors we discovered and which methods were able to most reliably discover each one. Of clear significance, is the increase in returned ''Random" (uninteresting) behaviors when observing Learned Embedding and ResNet approaches compared to Hand-Crafted Metrics. Coincidentally, ResNet18 achieves the lowest frequency of Random behaviors when using Random Search, showcasing the tendency for Novelty Search to deeply explore random behaviors that appear often in clustering.  

\subsection{Discovered Behaviors}
\label{appendix:discovered-behaviors}

\begin{figure*}[h]
     \centering
     \includegraphics[width=\linewidth]{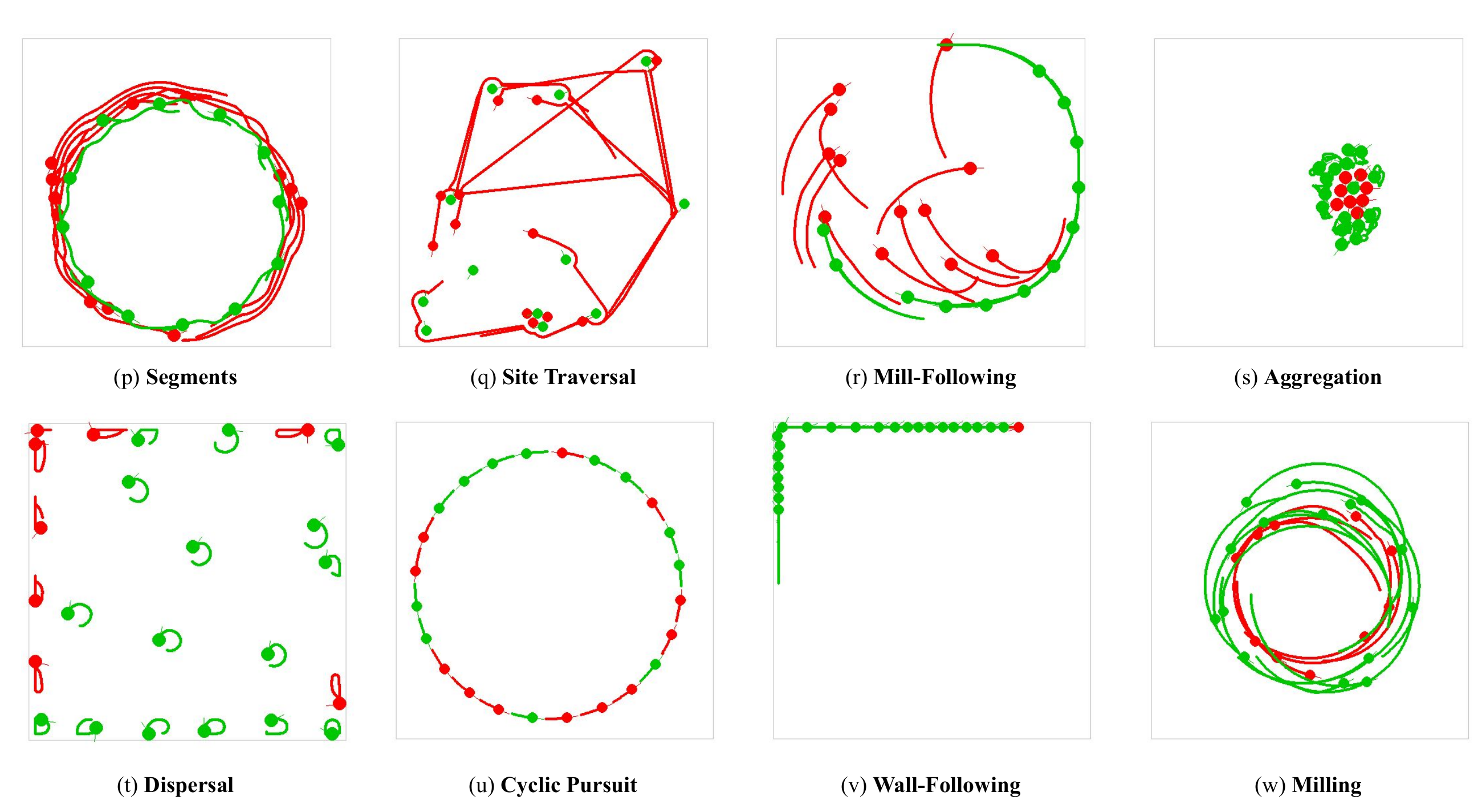}
     \caption{8 Additional Behaviors Reported by our experiments. Segments, Site Traversal, and Mill-Following have not been reported for agents of these capabilites before. The remaining behaviors (s-w) have been discovered by \cite{brown2018discovery} in prior work.}
    \label{fig:appendix-discovered}
\end{figure*}

Our work discovers 23 distinct emergent behaviors resulting from local interactions between 2 behavior-types of robots. In addition to the 15 presented in the main paper (Fig. \ref{fig:discovered}), we show an additional 8 in Fig. \ref{fig:appendix-discovered}. 

Because the classification of resulting emergent behaviors is subjective and based on the human's notion of novel and interesting, we include a short description of behaviors that guided our experiments in distinguishing between behaviors. We also include a vector from $\mathbb{H}_2$ that produces each behavior in the following paragraphs.

\subsubsection{Nested Cycles}
Agents of Type A form a circle around a circle of a smaller radius formed by agents of Type B. 
\textit{Controller:} [0.0, 0.5, 0.6, -0.1, 0.3, 0.7, 0.6, 0.0, $\frac{8}{24}$]

\subsubsection{Eye}
A small cyclic or milling pattern formed by Type A agents is encircled by Type B agents producing a large milling pattern.  
\textit{Controller:} [0.9, 1.0, 0.7, 0.7, 0.9, -0.7, 1.0, 0.8, $\frac{8}{24}$]

\subsubsection{Flower}
Type A agents aggregate in the center while Type B agents produce large ''pedals" or curves that fall away from the aggregating agents and then towards them again. 
\textit{Controller:} [0.7, 1.0, 0.4, 0.5, -0.9, -0.4, -0.3, 0.6, $\frac{12}{24}$]

\subsubsection{Spiral}
Type A agents form a cyclic pattern while Type B agents ''cling" tightly to a member of the Type A behavior, resulting in a cyclic pattern with many tails coming off of the central cycle.
\textit{Controller:} [0.1, 0.5, 0.6, -0.1, 0.3, 0.7, 0.4, -0.4, $\frac{8}{24}$]

\subsubsection{Nucleus}
Agents of Type A form a circle around Type B aggregating or tightly milling agents, resembling the nucleus of a cell. 
\textit{Controller:} [0.5, -0.7, 0.9, -0.5, 0.7, 1.0, 1.0, 0.5, $\frac{8}{24}$] 

\subsubsection{Flail}
Agents of Type A aggregate and bump into a line of follower agents (Type B). The continual aggregation bumping causes the line to spin around the aggregation agents, resembling a flail or chain being swung around a central pivot point.
\textit{Controller:} [-0.6, 1.0, 1.0, 0.4, 0.7, -0.6, 0.7, 1.0, $\frac{3}{24}$]

\subsubsection{Containment}
Agents of Type A attempt to disperse outwards from the center of the environment while agents of Type B attempt to ''fence-in" the dispersing agents by using a follower or cyclic pattern.
\textit{Controller:} [0.2, 0.7, -0.3, -0.1, 0.1, 0.9, 1.0, 0.8, $\frac{4}{24}$]

\subsubsection{Dipole}
Type A agents form a cyclic pattern around two opposing mills (Type B), where the mills are more attracted to the encircling agents than to each other, leading to a behavior where rotation occurs about two poles. 
\textit{Controller:} [1.0, -1.0, 0.7, 0.5, 0.9, 0.7, -1.0, -0.2, $\frac{12}{24}$]

\subsubsection{Snake}
Type A agents form a following pattern (Snake) where the leader agent seeks ''apples" (Agents of Type B). Type B agents are particularly good at dispersing/reversing away from detected snakes, leading to an exciting chase.
\textit{Controller:} [-0.7, 0.7, -0.4, -0.8, 0.8, 0.1, 0.2, 0.5, $\frac{1}{24}$]

\subsubsection{Hurricane}
Type A agents aggregate in the eye of a hurricane---A large milling pattern formed by agents of Type B.
\textit{Controller:} [-0.1, -0.2, 1.0, -1.0, 0.8, 0.9, 0.9, 1.0, $\frac{6}{24}$]

\subsubsection{Geometric Warp}
Type A agents disperse to corners and walls while a cyclic pattern (Type B) gets slowly warped due to the influence of the dispersing agents on the sensors of the cyclic agents, adding noise (warp) to the curve of the circle. 
\textit{Controller:} [-0.4, -1.0, -0.2, 0.9, -0.6, 0.7, 0.9, 1.0, $\frac{3}{24}$]

\subsubsection{Perimeter}
Type A agents mill or aggregate in the middle of the environment, while Type B agents wall follow at a distance. 
\textit{Controller:} [-0.9, -0.8, -0.8, -1., -0.6, -1., 0.9, -0.7, $\frac{6}{24}$]

\subsubsection{Mill + Disp.}
Type A agents mill in the center of the environment while Type B agents disperse away from the milling agents.
\textit{Controller:} [0.7, 1.0, 0.3, 0.4, 0.2, 0.7, -0.5, -0.1, $\frac{12}{24}$]

\begin{figure*}[t]
     \centering
     \includegraphics[width=0.8\textwidth]{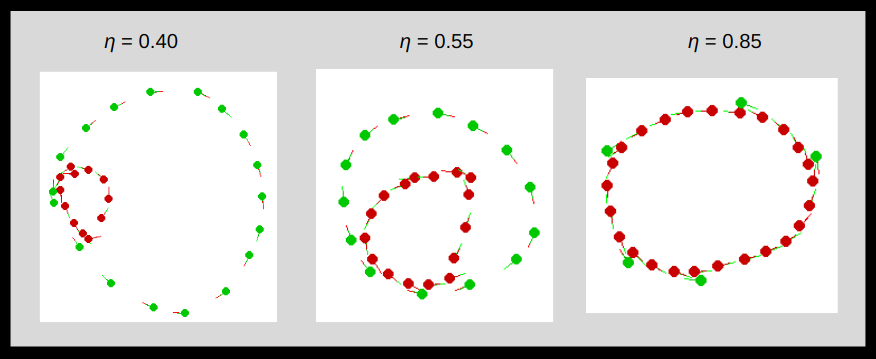}
     \caption{The effect of changing the population ratio ($\eta$) for in the following controller: [0.326, -0.579, 0.533, 0.472, 0.293, 0.424, 0.817, 0.795, $\eta$]. Images are captured at 1200 timesteps and are cropped for clarity.}
    \label{fig:phase-diagram-large}
\end{figure*}

\subsubsection{Agg. + Disp.}
Type A agents aggregate in the center of the environment while Type B agents disperse away from the aggregating agents.
\textit{Controller:} [0.1, 1.0, 0.3, 0.7, 0.2, 0.7, -0.5, -0.1, $\frac{12}{24}$]

\subsubsection{Cyclic + Disp.}
Type A agents form a cyclic pursuit pattern in the center of the environment while Type B agents disperse away from the cyclic agents.
\textit{Controller:} [0.6, 1.0, 0.4, 0.5, 0.2, 0.7, -0.5, -0.1, $\frac{12}{24}$]

\subsubsection{Segments}
Type A agents form a cyclic pattern and Type B agents form a cyclic pattern of similar radius but a much faster speed, causing a slowly rotating inner-cycle and a fast rotating outer cycle.
\textit{Controller:} [-0.9, 0.6, 0.9, 0.7, -0.4, 0.1, 0.6, 0.2, $\frac{12}{24}$]

\subsubsection{Site Traversal}
Type A agents remain static or nearly-static in their initial starting configuration while Type B agents dart from one Type A agent (site) to the next. The fast turning rate of Type B agents means that not all agents will take the same traversal path between sites, as shown in Fig. \ref{fig:appendix-discovered}q.
\textit{Controller:} [-0.9, 1.0, 1.0, 1.0, 0.1, -0.1, 0.0, 0.0, $\frac{12}{24}$] 

\subsubsection{Mill-Following}
Type A agents perform a milling pattern while Type B Agents follow the outside of the mill. When the milling agents see the trail of followers, they turn tightly inward, creating a crashing wave and the behavior repeats.
\textit{Controller:} [1.0, 0.9, 0.9, 0.5, 0.7, 0.5, 1.0, 1.0, $\frac{12}{24}$]

\subsubsection{Aggregation}
Type A and Type B agents attract each other and aggregate in the middle of the environment.
\textit{Controller:} [0.4, -0.7, 0.9, -0.5, 0.9, -0.4, 1.0, 0.4, $\frac{8}{24}$]

\subsubsection{Dispersal}
Type A and Type B agents repel each other and disperse outward from the middle of the environment.
\textit{Controller:} [-0.3, 0.1,  -0.4, -0.3, -0.3, 0., -0.2, -0.1, $\frac{8}{24}$]

\subsubsection{Cyclic Pursuit}
Type A and Type B agents form a perfect circle with evenly spaced agents forming the circumference.
\textit{Controller:} [-0.7, 0.3, 1.0, 1.0, -0.7, 0.3, 1.0, 1.0, $\frac{12}{24}$]

\subsubsection{Wall-Following}
Type A and Type B agents follow the 4 walls of the environment.
\textit{Controller:} [1.0, -0.1, -0.9, -1.0, 1.0, 0.6, -0.3, 0.9, $\frac{1}{24}$]

\subsubsection{Milling}
Type A and Type B agents rotate around a central pivot point but do not form an evenly spaced circle.
\textit{Controller:} [0.7, 1.0, 0.4, 0.5, 0.7, 0.9, 0.4, 0.5, $\frac{8}{24}$]

\subsection{Swarm Chemistry Implementation}
We seek to establish a baseline for automated behavior discovery by performing evolutionary search using a human as the fitness function instead of relying on a learned embedding. To achieve this, we draw inspiration from \textit{Swarm Chemistry}, which presents a framework for human-in-the-loop swarm evolution. \textit{Swarm Chemistry} presents a human user with an initial array of swarm behaviors. The human is tasked with the role of the \textit{alchemist}, and must select swarms for evolution which they find interesting/unique. These swarms are then randomly mutated and/or combined to form the next generation of swarms. Because the nature of the heterogeneous capability model presented in this paper differs fundamentally from that of \textit{Swarm Chemistry}, so too does the evolution pipeline. Regardless, the underlying principle of using a human as the fitness function for evolution remains unchanged.

We initialize the evolution pipeline by randomly generating 8 heterogeneous controllers. If the user wishes to replicate the starting conditions of the pipeline, they may set a random seed so that this operation becomes deterministic.

\begin{figure}[t]
     \centering
     \includegraphics[width=0.9\linewidth]{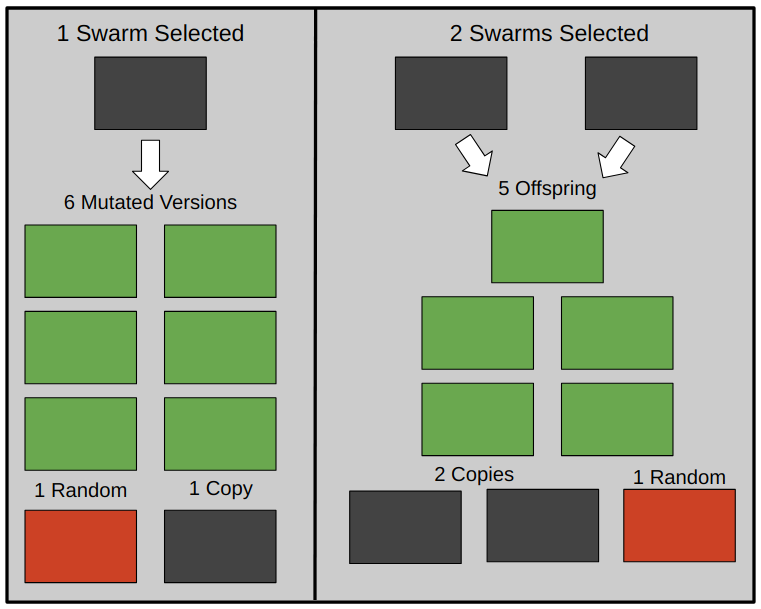}
     \caption{A diagram showing how the next generation of controllers is produced in our implementation of Swarm Chemistry. The process differs depending on the number of controllers the user selects.}
    \label{fig:1vs2clicks}
\end{figure}

The 8 controllers are simulated on a grid for the user. If the user finds a behavior interesting/unique, they may "save" it using a designated button. The user may save as many behaviors as they like each generation. The corresponding controller is saved to a file for later reference. The user may select between 1 and 2 swarms for evolution. When the user has selected the desired number of swarms, they press a button labeled "Advance" on the right side panel of the GUI. There are two other buttons labeled "Back" and "Skip", but their functionalities were not used in the context of this paper. If the user selects 1 swarm, the next generation of controllers will consist of 1 copy of the selected swarm's controller, 6 randomly mutated versions of the selected swarm's controller, and 1 randomly generated controller. If the user selects 2 swarms, the next generation of controllers will consist of the 2 controllers corresponding to the selected swarms, 1 randomly generated controller, and 5 \textit{offspring} of the 2 controllers corresponding to the selected swarms. Fig. \ref{fig:1vs2clicks} shows this process.

The random mutation operation applies the following operations to each element of the controller $c$ to produce a randomly mutated controller $c'$. Random crossover takes two controllers as input and produces an \textit{offspring} controller.

\subsection{Justification for Population Ratio Value}

The \textit{population ratio} parameter ($\eta$) in heterogeneous controllers dictates the ratio between the populations of the two controller subspecies. For some controllers, the behavior of a swarm may remain fundamentally the same despite large changes to $\eta$, as shown in Fig. \ref{fig:phase-diagram-large}.

Conversely, relatively small changes to $\eta$ can lead to large and unpredictable changes in the behavior of the swarm. Ultimately, the effect that changing $\eta$ will have on the behavior of the swarm depends on the rest of the parameters in the controller. Restricting $\eta$ to a constant such as $k=0.5$ would limit our ability to explore the entirety of the behavior space because some behaviors would be impossible to simulate. For example, if $\eta$ were restricted to $0.5$, it would be impossible to simulate the \textit{Snake} behavior shown in Fig. \ref{fig:discovered}. As a means to the end of exploring the largest number of distinct heterogeneous swarm behaviors, we included the population ratio in the heterogeneous controller.

\addtolength{\textheight}{-12cm}   

\end{document}